\documentclass[sigconf]{acmart} 

\usepackage[a-1b]{pdfx}

\usepackage{graphicx}
\usepackage{balance}
\usepackage{comment}
\usepackage{amsfonts}
\usepackage{color,soul}
\usepackage{hyperref}
\usepackage{subcaption}
\usepackage{xspace}
\usepackage{boxedminipage}
\usepackage{diagbox}
\usepackage{multirow}
\usepackage{paralist}
\usepackage{amsmath}
\usepackage{enumitem}
\usepackage{array}
\usepackage{mdframed}
\usepackage[normalem]{ulem}
\usepackage{tabularx}
\usepackage{pgf,tikz}
\usepackage{pgfplots}
\usepackage{mathrsfs}
\usepackage{diagbox}
\usepackage{footnote}
\usepackage{booktabs}
\usepackage{wrapfig}
\usepackage{url}
\usepackage{amsmath}
\usepackage{newtxmath}
\usepackage{arydshln}
\usepackage[normalem]{ulem}
\usepackage{xcolor}
\usepackage{colortbl}
\usepackage{ctable}
\usepackage{textcomp}
\usepackage{accents}
\useunder{\uline}{\ul}{}

\usepackage{cleveref}[2012/02/15]
\crefformat{footnote}{#2\footnotemark[#1]#3}

\makeatletter
\DeclareRobustCommand{\iscircle}{\mathord{\mathpalette\is@circle\relax}}
\newcommand\is@circle[2]{%
  \begingroup
  \sbox\z@{\raisebox{\depth}{$\m@th#1\bigcirc$}}%
  \sbox\tw@{$#1\square$}%
  \resizebox{!}{\ht\tw@}{\usebox{\z@}}%
  \endgroup
}
\makeatother

\usepackage[linesnumbered,ruled,vlined]{algorithm2e}

 \newcolumntype{L}[1]{>{\raggedright\arraybackslash}p{#1}}
\SetCommentSty{algcmmt}
\SetKwInput{KwInput}{Input}                
\SetKwInput{KwOutput}{Output}

\newcommand{\algname}{SCStory}

\AtBeginDocument{%
  \providecommand\BibTeX{{%
    \normalfont B\kern-0.5em{\scshape i\kern-0.25em b}\kern-0.8em\TeX}}}

\copyrightyear{2023}
\acmYear{2023}
\setcopyright{acmlicensed}\acmConference[WWW '23]{Proceedings of the ACM
Web Conference 2023}{May 1--5, 2023}{Austin, TX, USA}
\acmBooktitle{Proceedings of the ACM Web Conference 2023 (WWW '23), May
1--5, 2023, Austin, TX, USA}
\acmPrice{15.00}
\acmDOI{10.1145/3543507.3583507}
\acmISBN{978-1-4503-9416-1/23/04}

\begin{document}

\title{\algname{}: Self-supervised and Continual Online Story Discovery}


\author{Susik Yoon}
\affiliation{%
  \institution{UIUC}
}
\email{susik@illinois.edu}

\author{Yu Meng}
\affiliation{%
  \institution{UIUC}
}
\email{yumeng5@illinois.edu}

\author{Dongha Lee}
\affiliation{%
  \institution{Yonsei University}
}
\email{donalee@yonsei.ac.kr}

\author{Jiawei Han}
\affiliation{%
  \institution{UIUC}
}
\email{hanj@illinois.edu}


\begin{abstract}
We present a framework \algname{} for online story discovery, that helps people digest rapidly published news article streams in real-time without human annotations. To organize news article streams into stories, existing approaches directly encode the articles and cluster them based on representation similarity. However, these methods yield noisy and inaccurate story discovery results because the generic article embeddings do not effectively reflect the story-indicative semantics in an article and cannot adapt to the rapidly evolving news article streams. \algname{} employs self-supervised and continual learning with a novel idea of story-indicative adaptive modeling of news article streams. With a lightweight hierarchical embedding module that first learns sentence representations and then article representations, \algname{} identifies story-relevant information of news articles and uses them to discover stories. The embedding module is continuously updated to adapt to evolving news streams with a contrastive learning objective, backed up by two unique techniques, confidence-aware memory replay and prioritized-augmentation, employed for label absence and data scarcity problems. Thorough experiments on real and the latest news data sets demonstrate that \algname{} outperforms existing state-of-the-art algorithms for unsupervised online story discovery.
\end{abstract}



\maketitle

\section{Introduction}
These days, news articles covering real-time events are massively published through online platforms. Real-time discovery of news stories with articles under unique themes brings huge benefits not only for individuals to follow emerging news stories but also for organizations to make strategic decisions. This calls for practical solutions for \textit{online news story discovery}, thereby enabling downstream tasks such as summarization and recommendation\,\citep{topicexpan, story_summary, news_recom}, event detection and tracking\,\citep{cep, mdual, arcus}, and curation\,(e.g., Google News).

Existing studies for online news story discovery embed articles and incrementally cluster them into stories. Some of them utilize external knowledge (e.g., entity or story labels) to train embedding and clustering models\,\citep{miranda, storyforest, saravan}. Another line of efforts adopts an \emph{unsupervised} approach, which is more practical in an online scenario, with symbolic- or graph-based embedding based on keywords statistics and online clustering\,\citep{constream, newslens, staykovski}. These studies, however, have fundamental limitations in capturing \emph{diversified} and \emph{evolving} context of news article streams, as they use static and explicit features for embedding and clustering. News articles often contain diverse descriptions depending on the reporters' perspectives and writing styles, some of which may not directly indicate the major theme of the stories. This makes distinguishing stories more challenging because the same story may be reported with diverse themes by different media sources, whereas different stories from the same media source may contain overlapping themes. Moreover, the themes will gradually change over time as new articles with fresh stories are rapidly published. 

\begin{figure*}
    \centering
    \includegraphics[width=\textwidth]{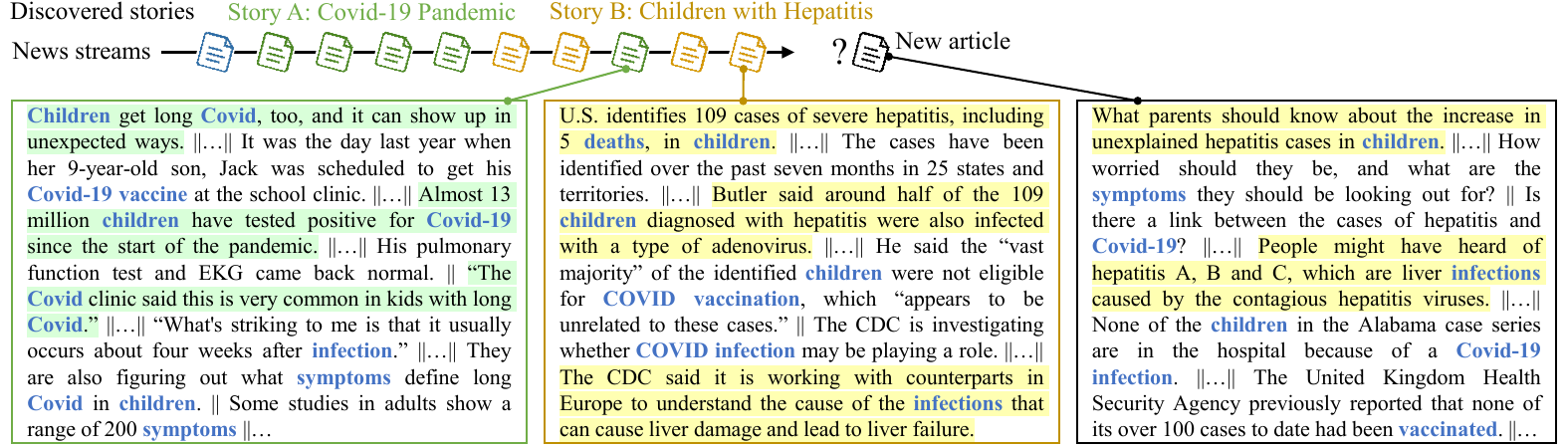}
    \vspace{-0.7cm}
    \caption{The articles contain common keywords (marked in {\textbf{\color{blue} blue}}) and too similar/specific semantics. The story-indicative parts (highlighted in \colorbox{green!15}{green} for Story A and in \colorbox{yellow!30}{yellow} for Story B) are helpful to clearly distinguish articles in different stories.}
    \label{fig:motivation}
    \vspace{-0.45cm}
\end{figure*}

For instance, in Figure \ref{fig:motivation}, news articles of two stories, A: \texttt{Covid-19 Pandemic} and B: \texttt{Children with Hepatitis}, are being published. When a new article is given, existing approaches indiscriminately encode the whole content based on keywords statistics or a static embedding model without considering its story-distinctiveness. While the two stories are clearly different, their articles contain common keywords (e.g., `\emph{children}', `\emph{covid}', and `\emph{infection}') and descriptions with too similar (e.g., `\emph{many children have caught ...}') or too specific (e.g., `\emph{her 9-year-old son, Jack, ...}') semantics. The keywords and semantics will also dynamically change as stories evolve, making it more difficult to cluster articles into correct stories.

To tackle this problem and enable more effective online story discovery, we propose a novel idea called \textbf{story-indicative adaptive modeling} of news article streams. In Figure \ref{fig:motivation}, the sentences highlighted in each article are more specifically relevant to the corresponding story than to another story. If one can identify and focus on such story-relevant information in each article in the context of concurrent stories, it becomes much clearer that the new article should fall in Story B. However, there are non-trivial technical challenges to achieve this goal: (1) Story-indicative semantics need to be identified from the unstructured and diverse contents (i.e., \emph{diversified local context}). (2) The ever-changing distinctive themes shared among news articles in concurrent stories must be considered (i.e., \emph{evolving global context}). (3) It is almost impossible to obtain human annotations due to the rapidly and massively published news articles (i.e., \emph{no supervision}). (4) The model should be scalable and adaptable to such rapid and massive news article streams (i.e., \emph{efficient adaptation}).

To this end, we design \textbf{\algname{}}, a framework for \emph{\textbf{S}elf-supervised and \textbf{C}ontinual online \textbf{Story} discovery}.
\algname{} features a hierarchical embedding learning where quality sentence representations of an article are first obtained from a pretrained sentence encoder\,\citep{sentencebert}, and then a lightweight article embedding model fine-tunes the sentence representations further and derives the article representation promoting story-indicative semantics with the diversified local context. Over news article streams, similarities between the articles are continuously inspected to cluster them into distinctive stories. In the meantime, the model is updated by contrastive learning with confident article-story assignments as self-supervision to cater to the evolving global context. The update is accompanied by two unique techniques, \emph{confidence-aware memory replay} to overcome label absence and \emph{prioritized-augmentation} to overcome data scarcity, for robust knowledge preservation over time.

In brief, the contributions of \algname{} are:
\begin{itemize}[leftmargin=12pt, noitemsep]
\item To the best of our knowledge, this is the first work to apply \emph{self-supervised} and \emph{continual} learning for unsupervised online story discovery from text-rich article streams.
\item We propose \emph{a novel idea of story-indicative adaptive modeling} to capture only the story-relevant information from dynamically evolving articles and stories.
\item We propose \emph{a lightweight story-indicative embedding model} to realize the idea, trained by contrastive learning with confidence-aware memory replay and prioritized-augmentation. The source codes of \algname{} are publicly available\,\footnote{\url{https://github.com/cliveyn/SCStory}}.
\item We demonstrate the performance improvement of \algname{} over state-of-the-art unsupervised online story discovery methods through extensive evaluations with real news data sets and an in-depth case study with the latest news stories.
\end{itemize}
\section{Preliminaries}
\begin{figure*}
    \centering
    \includegraphics[width=\textwidth]{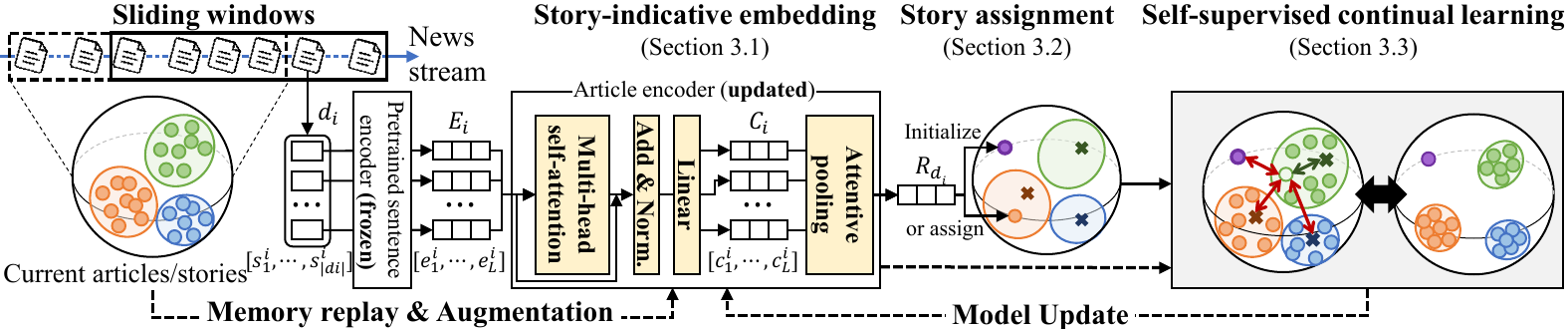}
    \vspace{-0.5cm}
    \caption{In every sliding window of an article stream, \textbf{first}, \algname{} learns story-indicative article representations by modeling the correlations between initial sentence representations from a PSE. \textbf{Second}, \algname{} assigns each new article to the most confident current story or forms a new story. \textbf{Lastly}, \algname{} updates the story-indicative embedding module based on the contrastive loss with confidence-based memory replay and prioritized-augmentation (the base PSE is frozen).}
    \vspace{-0.2cm}
    \label{fig:overview}
\end{figure*}
\subsection{Problem Setting}
A news article\,(or simply \emph{article}) $d$ is a set of sentences $[s_1, s_2, \ldots, s_{|d|}]$ and a news story\,(or simply \emph{story}) $S$ is a set of articles $[d_1, d_2, \ldots, d_{|S|}]$ consecutively published and describing a relevant event with a unique theme (e.g., \texttt{Russia-Ukraine War}). The task of unsupervised online story discovery is to incrementally cluster each new article in an article stream $\mathbb{D} = [\ldots, d_{i-1}, d_{i}, d_{i+1}, \ldots]$ into one of the discovered (or new) stories $\mathbb{S} = [\ldots, S_{j-1}, S_{j}, S_{j+1}, \ldots]$ without supervision.
To work with the unbounded news streams, we adopt a concept of \emph{sliding window} $\mathbb{W} \subseteq \mathbb{D}$, widely used for mining data streams\,\citep{stream_clus, newslens, nets, stare}. The window size and the slide size respectively represent a time scope of interest for ongoing stories and its update frequency (e.g., a sliding window of 7 days slid by a day indicates the articles published in the last week and updated daily). Practically, we assume that each article belongs to a single story (i.e., $\sum_{S_j\in\mathbb{S}}|d_i\!\in\!S_j|\!=\!1,\!\forall d_i\!\in\!\mathbb{D}$) and a story is active if any article in the window $\mathbb{W}$ falls in the story (i.e., $\sum_{d_i \in \mathbb{W}}|d_i \in S_j|\!\geq\!1,\!\forall S_j\!\in\!\mathbb{S}_{\mathbb{W}}$ where $\mathbb{S}_{\mathbb{W}}$ are the ongoing stories in $\mathbb{W}$)\footnote{It is natural that articles in different stories are rarely overlapped in news outlets and a story is regarded as inactive when there have been no updates for a long time.}.






\subsection{Overview of Article Embedding Learning}
Pretrained language models (PLMs)~\citep{bert,longformer,cocolm} have demonstrated remarkable embedding power to encode textual semantics into contextualized representations.
The straightforward deployment of PLMs for sequence-level tasks directly learns sequence representations over all tokens in the input sequence. 
In this work, however, we follow a \emph{hierarchical} approach, to first get individual sentence representations in an article using pretrained sentence encoders (PSEs)~\citep{sentencebert,sentencet5} and then derive a story-indicative article representation from them via a lightweight attention-based encoder.

Such a hierarchical approach has the following benefits: 
(1) Articles are usually long; embedding them directly with PLMs, mostly based on Transformers~\citep{transformer}, causes large computation overhead due to the quadratic scaling of the self-attention mechanism to the long input sequence length.
Breaking down an article into sentences improves efficiency by allowing parallel encoding of short sentences. 
(2) The major theme of an article usually concentrates on a few sentences with distinct semantics. Explicitly investigating sentences facilitates the discovery of such story-indicative sentences, which are expected to play an essential role in embedding articles.
(3) Under the unsupervised setting, there lacks explicit annotation to directly train an embedding model from scratch. Leveraging the readily available quality sentence representations from PSEs allows the model to derive good initial article representations, based on which the model can gradually refine via a self-supervised objective.

\section{The Proposed Framework: \algname{}}
Figure \ref{fig:overview} shows the overall procedure of \algname{}, to realize story-indicative adaptive modeling for unsupervised online story discovery. In brief, \algname{} employs a lightweight article encoder on the top of a frozen PSE for extracting the contextual semantics of articles (Section \ref{sec:self_attentive_embedding}). While each new article in a sliding window is embedded by the encoder and clustered to stories incrementally (Section \ref{sec:story_assignment}), the encoder itself is updated with self-supervised continual learning to make articles in different stories more discriminative (Section \ref{sec:self_supervised_continaul_learning}). The following sections detail each step.



\subsection{Story-indicative Embedding}
\label{sec:self_attentive_embedding}
\textbf{\uline{Initial sentence embedding.} } Each sentence in a new article $d_i\!=\![s_1^i,\ldots,s_{|d_i|}^i]$ is fed into a PSE to get initial sentence representations. Given a PSE, and a maximum number $L$ of sentences, initial sentence embedding (ISE) of $d_i$ is formulated as:\!
\begin{equation}
\label{eq:init_sent_repr}
\begin{split}
E_i &=\text{ISE}(d_i) = \text{TP}([\text{PSE}(s_1^i),\ldots,\text{PSE}(s_{|d_i|}^i)],L)\\ 
&= [e_1^i,\ldots,e_L^i] \in \mathbb{R}^{L \times h_{e}},
\end{split}
\end{equation}
where $h_{e}$ is the output dimensionality of PSE, and TP$(\cdot,L)$ truncates or pads an input sequence to match the length $L$, which can be set large enough (e.g., 50 sentences) to cover the most articles. The padded part will be ignored with masking.

\noindent
\textbf{\uline{Story-indicative article encoder.} }
Given initial representations of sentences in an article, a story-indicative article encoder $\mathcal{M}$ derives the unique semantics of the article in the context of its story by investigating the relationships between sentences, through a \emph{multi-head self-attention} block and an \emph{attentive pooling} block.

\noindent
\textbf{\uline{Story-indicative sentence embedding.} } In the first block, \algname{} adopts a sentence-level multi-head self-attention (MHS) mechanism\,\citep{transformer}, inspired by the token-level MHS commonly used in Transformers. The general idea is to automatically learn story-indicative sentence representations by modeling the interdependent correlations among sentences. Specifically, \algname{} extracts contextual features from different subspaces in $E_i$:
\begin{equation}
    \text{MHS}(E_i) = [H_1, \ldots, H_n]W^O \text{, where}
\end{equation}
$H_j\!=\!\text{softmax}(E_iW_j^Q(E_iW_j^K)^\top/\sqrt{h_{e}/n})E_iW_j^V$, $n$ is the number of heads, $h_e$ is the hidden dimensionality of MHS, and $W^O\!\in\!\mathbb{R}^{h_{e} \times h_{e}}$, $W_j^Q,\!W_j^K,\!W_j^V\!\in\!\mathbb{R}^{h_{e} \times \frac{h_{e}}{n}}$ are learnable parameters. The output of MHS is further processed with layer normalization (LN) followed by a residual connection, and finally a linear layer. Specifically, given initial sentence representations $E_i$ of an article $d_i$, story-indicative sentence representations $C_i$ of $d_i$ are formulated as:
    \begin{equation}
    \label{eq:sentence_repr}
        \begin{split}
            C_i &= \text{tanh}(\text{LN}(\text{MHS}(E_i\text) + E_i)W_c + b_c) \\
           &= [c_1^i,\ldots,c_L^i] \in \mathbb{R}^{L \times h_{c}},
        \end{split}
    \end{equation}
where $W_c, b_c$ are learnable parameters and $h_c$ is the output dimensionality of the linear layer.

\noindent
\textbf{\uline{Story-indicative article embedding.} }
In the attentive pooling block, \algname{} learns attention weights for weighted pooling of $C_i$ to get a single story-indicative article representation. The attention weights are derived through an attention layer which reflects how story-relevant each sentence is. Specifically, given $C_i= [c_1^i,\ldots,c_L^i]$ of an article $d_i$, a story-indicative article representation $R_{d_i}$ is:
\begin{equation}
\label{eq:article_rpr}
    R_{d_i} = \sum_{j=1}^L \alpha_j c_j^i \in \mathbb{R}^{h_c}, \text{where}
\end{equation}
$\alpha_j =[$softmax$($tanh$(C_i W_a + w_a)^{\top}V_a)]_j$ is an attention weight for the sentence $j$ and $W_a, w_a, V_a$ are learnable parameters.

In a nutshell, the story-indicative article encoder $\mathcal{M}$ encodes the initial sentence representations of an article $d_i$ into a story-indicative article representation, i.e., $R_{d_i} = \mathcal{M}(\text{ISE}(d_i))$. All the learnable parameters in $\mathcal{M}$ are trained under the learning mechanism to be introduced in Section \ref{sec:self_supervised_continaul_learning}.

\subsection{Story Assignment}
\label{sec:story_assignment}
\textbf{\uline{Dynamic story embedding.} }\algname{} dynamically represents a story by using the articles in the story especially in the context of the current window to reflect the latest shared themes of the corresponding articles. 
Given a current window $\mathbb{W}$, and a set $\mathbb{S}_{\mathbb{W}}$ of current stories in $\mathbb{W}$, a representation $R_{S_j}$ of a story $S_j \in \mathbb{S}_{\mathbb{W}}$ is defined as the mean of the story-indicative article representations:
\begin{equation}
\label{eq:story_rpr}
R_{S_j} = \frac{1}{|S_j \cap \mathbb{W}|}\sum_{d_i \in S_j \cap \mathbb{W}}\mathcal{M}(\text{ISE}(d_i)).
\end{equation}

\noindent
\textbf{\uline{Article-story confidence.} } Then, \algname{} decides how confidently a new article can be assigned to a story by comparing their representation similarity. Given a new article $d_i$ and a target story $S_j$, an article-story confidence score is computed as:\!
    \begin{equation}
    conf_{d_i|S_j} = \text{cos}(R_{d_i}, R_{S_j}),
    \end{equation}
where cos($\cdot$,$\cdot$) is a cosine similarity and $R_{d_i}$ and $R_{S_j}$ are derived by Equations \ref{eq:article_rpr} and \ref{eq:story_rpr}, respectively.

If the maximum confidence score of a new article to one of the current stories exceeds a threshold (i.e., $max(\{conf_{d_i|S_j}|S_j \in \mathbb{S}_{\mathbb{W}}\}) > \delta$), the article is added to the corresponding story, or otherwise, it becomes a seed article for a new story. For the threshold which controls the granularity of a story, we use $\delta=0.5$ as a default value which is a neutral similarity by definition and also demonstrated as a practical boundary between positive and negative articles in real news articles (refer to Section \ref{exp:analysis} for more details). Note that it can be adjusted dynamically or set manually based on domain knowledge or application requirements.

\subsection{Self-supervised Continual Learning}
\label{sec:self_supervised_continaul_learning}
In every sliding window, \algname{} continually optimizes $\mathcal{M}$ towards two goals: \emph{contrastive learning with self-supervision} and \emph{continual learning with memory replay}, where the former promotes articles to be more closely clustered towards their stories (i.e., more story-indicative) and the latter regularizes the contrast to be temporally consistent throughout the evolving article stream (i.e., more robust to catastrophic forgetting\,\citep{kirkpatrick2017overcoming}). However, the \emph{label absence} and the \emph{data scarcity} problems that naturally arise in an unsupervised online scenario make these goals challenging. 

To this end, \algname{} \emph{samples} and \emph{augments} training batch of articles from the current window and exploits their cluster (i.e., story) membership confidence as a means of self-supervision. Then, $\mathcal{M}$ is trained to represent articles being closer to their clusters (i.e., to positive articles) while being further from different clusters (i.e., from negative articles). The corresponding two techniques for preparing training samples, \emph{confidence-aware memory replay} and \emph{prioritized-augmentation}, and the consequent contrastive learning mechanism are explained in detail as follows.

\noindent
\textbf{\uline{Confidence-aware memory replay.} } The most straightforward way to adapt $\mathcal{M}$ over an article stream is to train it incrementally with new articles in a new slide. However, only focusing on the new articles is impractical not only because their true story labels are unknown but also because their fluctuating distributions can distort the consistency of $\mathcal{M}$. Instead, \algname{} adopts a philosophy of \emph{memory replay}\,\citep{lifelongLM, continualLM}, which retrieves informative previous samples for robust training, and implements it with a unique sampling and pseudo-labeling strategy to overcome label absence. 

Specifically, \algname{} sets the scope of memory replay to the current window $\mathbb{W}$ which caches articles within a recent time frame and samples articles with confident pseudo-story labels. This preserves previously learned knowledge within the latest temporal context of interest and helps keep track of ongoing discovered stories. More formally, an article $d$ is sampled from $\mathbb{W}$ and gets a pseudo-story label $S$ with the sampling probability $p(d|S)$ which is proportional to the confidence score $conf_{d|S}$ of the article and a story in $\mathbb{S}_{\mathbb{W}}$. Thus, the more similar a story-indicative article representation and a story representation, the higher probability that the article is sampled with the corresponding pseudo-story label:
\begin{equation}
 p(d | S) \propto conf_{d|S} = cos(R_{d},R_{S}) \text{ for } d \in \mathbb{W}, S \in \mathbb{S}_{\mathbb{W}}.
\end{equation}

\noindent
\textbf{\uline{Prioritized-augmentation.} } Because stories naturally have different lifespans, some stories in the current window may be underrepresented as their articles are sparsely distributed over time, especially when they are newly formed, gradually progressed, or recently expired. This can degrade the consistency and generalization capabilities of $\mathcal{M}$. \algname{} tackles the \emph{data scarsity} problem with a unique article augmentation strategy.

The augmentation for linearly interpolating input samples and their labels, such as Mixup\,\cite{zhang2018mixup}, has proven to be effective for self- or semi-supervised learning and has been widely studied in relevant NLP tasks such as classification and language model pretraining\,\cite{chen2020mixtext,guo2019augmenting,sun2020mixup}. We introduce a novel sentence-level article augmentation technique specifically designed for the story discovery task, that concatenates the top half sentences in $d_i$ and the bottom half sentences in $d_j$, respectively prioritized by their intra-article importances in deriving story-indicative article representations. Specifically, given two articles $d_i$ and $d_j$ assigned to the same story, the prioritized-augmentation p-aug($\cdot$,$\cdot$) is formulated as follow:
\begin{equation}
\begin{split}
&\text{p-aug}(d_i,d_j)  \\
&=\text{concat}([s_1^{i\prime},\ldots,s_{\frac{|d_i|}{2}}^{i\prime}|s_k^{i\prime} \in d_i],[s_{\frac{|d_j|}{2}}^{j\prime},\ldots,s_{|d_j|}^{j\prime}|s_k^{j\prime} \in d_j]),
\end{split}
\end{equation}
where $s_k^{i\prime}$ and $s_k^{j\prime}$ are the sentences respectively in $d_i$ and $d_j$ and ordered by the normalized weights of the multi-head self-attention layer MHS in Equation \ref{eq:sentence_repr} (i.e., a weight for a sentence $s_k$ is calculated by sum$(W_O|_{k,:})/$sum$(W_O|_{:,:})$ where $W_O|_{\cdot,\cdot}$ denotes the indexing at a matrix $W_O$)\footnote{More effective than the final pooling weights in Equation \ref{eq:article_rpr} which does not incorporate pairwise relationships between sentences (refer to ablation study in Section \ref{exp:analysis}).}. This ensures that the augmented article contains both the story-indicative and the story-irrelevant information for the story $S$, so as to help generalization of $\mathcal{M}$ with diverse samples and overcome the data scarcity problem. 

\noindent
\textbf{\uline{Training samples.} } Finally, a set of training samples $\mathbb{B}$ is constructed by sampling documents from $\mathbb{W}$ with a sampling probability $p(\cdot|\cdot)$ for the confidence-aware memory replay and by augmenting documents with p-aug$(\cdot,\cdot)$ for the prioritized-augmentation:
\begin{equation}
\label{eq:training_samples}
\begin{split}
    \mathbb{B} &= \{(d,S)|d \in \mathbb{W}, d \sim p(d | S) \propto conf_{d|S}, S \in \mathbb{S}_{\mathbb{W}}\} \\
    &\cup \{(\text{p-aug}(d_i,d_j),S)|d_i,d_j \in S \cap \mathbb{W}, S \in \mathbb{S}_{\mathbb{W}}\}, \\ 
\end{split}
\end{equation}

\noindent
\textbf{\uline{Contrastive loss.} } Then, we use the contrastive loss\,\citep{infonce} calculated with article-story pairs $(d,S) \in \mathbb{B}$ for training $\mathcal{M}$:
\begin{equation}
\label{eq:loss}
    \mathcal{L}_{cts}\!=\!-\!\sum_{(d,S) \in \mathbb{B}}\text{log}\frac{e^{\text{cos}({R_{d},R_{S})}/\tau}}{\sum_{S' \in \mathbb{S}_{\mathbb{W}}}e^{\text{cos}({R_{d},R_{S'})}/\tau}}.
\end{equation}
Minimizing $\mathcal{L}_{cts}$ helps positive articles in the same story align closer (i.e., by maximizing the numerator) while negative articles in different stories are pushed away (i.e., by minimizing the denominator) in the embedding space. In other words, it optimizes \algname{} to improve the alignment and uniformity properties\,\citep{align_uniform} of the embedded articles, which will be detailed in Section \ref{sec:theo_analysis}.

The pseudocode of \algname{} is provided in Algorithm 1. For the very first window in a cold-start scenario, \algname{} uses mean pooling of initial sentence representations for embedding articles (a pre-trained story-indicative article encoder, if available, can be used). Then, \algname{} applies a cluster seed selection (e.g., k-means++\,\citep{kmeans++}) only for the story initialization purpose.

\begin{algorithm}[!t]
\label{alg:overall}
\small
\caption{A pseudocode of \algname{}}
\DontPrintSemicolon
\KwInput{an article stream $\mathbb{D}$}
\KwOutput{a set $\mathbb{S}_{\mathbb{W}}$ of stories in sliding window $\mathbb{W}$}
$\mathbb{S}_{\mathbb{W}} \leftarrow$ get initial stories from the first window\; 
\For{every sliding window $\mathbb{W}$ from $\mathbb{D}$}
{
    \For{new articles $d_i$ in $\mathbb{W}$}
    {
        $E_i \leftarrow \text{ISE}(d_i)$ with a PSE (Equation \ref{eq:init_sent_repr})\;
        $R_{d_i} \leftarrow \mathcal{M}(E_i)$ (Equation \ref{eq:article_rpr})\;
        \uIf{$max(\{conf_{d_i|S_j}|S_j \in \mathbb{S}_{\mathbb{W}}\}) \geq \delta $}
        {
            Assign $d_i$ to the corresponding $S_j$\;
        }\Else{
            $\mathbb{S}_{\mathbb{W}} \leftarrow$ add a new story $S = [d_i]$\;
        }
    }
    \For{epoch in epochs}
    {
        \For{itr in iterations}
        {
        $\mathbb{B} \leftarrow$ construct training samples (Equation \ref{eq:training_samples})\; 
        $\mathcal{L}_{cts} \leftarrow$ contrastive loss with $\mathbb{B}$ (Equation \ref{eq:loss})\;
        $\mathcal{M} \leftarrow$ update with $\mathcal{L}_{cts}$\;
        }
    }
    Report $\mathbb{S}_{\mathbb{W}}$;
}
\end{algorithm}
\newlength{\oldtextfloatsep}\setlength{\oldtextfloatsep}{\textfloatsep}
\setlength{\textfloatsep}{10pt}

\begin{table*}[!t]
    \caption{Online story discovery performance evaluation results (the average and standard deviation of scores from five runs are reported for \algname{}). The highest and second highest scores in each metric are bolded and underlined, respectively.}
    \vspace{-0.3cm}
    \label{tbl:overall_accuracy}
    \small
\centering
\setlength{\tabcolsep}{4.8pt}
\begin{tabular}{l|ccc|ccc|ccc}
\toprule
              & \multicolumn{3}{c}{News14} & \multicolumn{3}{c}{WCEP18} & \multicolumn{3}{c}{WCEP19} \\
              & $B^{3}$-F1   & AMI    & ARI    & $B^{3}$-F1 & AMI   & ARI   & $B^{3}$-F1 & AMI   & ARI   \\ \hline
ConStream{ \,\citep{constream}}     & 0.314        & 0.128  & 0.069  & 0.408      & 0.444 & 0.222 & 0.400      & 0.497 & 0.292 \\
NewsLens{ \,\citep{newslens}}      & 0.481        & 0.309  & 0.077  & 0.527      & 0.490 & 0.117 & 0.554      & 0.529 & 0.141 \\
BatClus{ \,\citep{miranda}}       & 0.706        & 0.726  & 0.572  & 0.694      & 0.786 & 0.571 & 0.698      & 0.791 & 0.574 \\
DenSps{ \,\citep{staykovski}}        & 0.669        & 0.602  & 0.358  & 0.697      & 0.759 & 0.487 & 0.701      & 0.765 & 0.487 \\ \hline 

ConStream+ST5 & 0.388        & 0.282  & 0.139  & 0.707      & 0.788 & 0.660 & 0.705      & 0.793 & \uline{0.669} \\
NewsLens+ST5  & 0.622        & 0.539  & 0.261  & 0.630      & 0.688 & 0.332 & 0.649      & 0.711 & 0.370 \\

BatClus+ST5   & 0.732        & 0.753  & 0.617  & 0.710      & 0.798 & 0.629 & 0.717      & 0.805 & 0.644 \\
DenSps+ST5    & 0.684        & 0.631  & 0.415  & 0.735      & 0.798 & 0.582 & 0.704      & 0.782 & 0.537 \\
\textbf{\algname{}+ST5}
&  \renewcommand{\arraystretch}{0.8} \begin{tabular}[c]{@{}c@{}} \uline{0.830} \\ {\scriptsize (0.0014)}\end{tabular}
&  {\renewcommand{\arraystretch}{0.8} \begin{tabular}[c]{@{}c@{}} \uline{0.847} \\ {\scriptsize (0.0016)}\end{tabular}}       
&  {\renewcommand{\arraystretch}{0.8} \begin{tabular}[c]{@{}c@{}} \uline{0.756} \\ {\scriptsize (0.0034)}\end{tabular}}        
&  {\renewcommand{\arraystretch}{0.8} \begin{tabular}[c]{@{}c@{}} \uline{0.794} \\ {\scriptsize (0.0014)}\end{tabular}}            
&  {\renewcommand{\arraystretch}{0.8} \begin{tabular}[c]{@{}c@{}} \uline{0.851}\\ {\scriptsize (0.0012)}\end{tabular}}       
&  {\renewcommand{\arraystretch}{0.8} \begin{tabular}[c]{@{}c@{}} \uline{0.720} \\ {\scriptsize (0.0021)}\end{tabular}}       
&  {\renewcommand{\arraystretch}{0.8} \begin{tabular}[c]{@{}c@{}} 0.718 \\ {\scriptsize (0.0054)}\end{tabular}}            
&  {\renewcommand{\arraystretch}{0.8} \begin{tabular}[c]{@{}c@{}} 0.803 \\ {\scriptsize (0.0026)}\end{tabular}}       
&  {\renewcommand{\arraystretch}{0.8} \begin{tabular}[c]{@{}c@{}} 0.633 \\ {\scriptsize (0.0045)}\end{tabular}}
\\ \hline 

ConStream+SBERT  & 0.434        & 0.413  & 0.276  & 0.701      & 0.784 & 0.657 & 0.704      & 0.795 & 0.667 \\
NewsLens+SBERT   & 0.749        & 0.718  & 0.564  & 0.767      & 0.823 & 0.631 & \uline{0.784}      & \uline{0.837} & 0.664 \\
BatClus+SBERT    & 0.764        & 0.785  & 0.648  & 0.751      & 0.835 & 0.656 & 0.759      & \uline{0.837} & 0.657 \\ 
DenSps+SBERT     & 0.750        & 0.720  & 0.567  & 0.754      & 0.824 & 0.642 & 0.762     & 0.830 & 0.660 \\ 
\textbf{\algname{}+SBERT}    
&  {\renewcommand{\arraystretch}{0.8} \begin{tabular}[c]{@{}c@{}}\textbf{0.872}\\ {\scriptsize (0.0014)}\end{tabular}}
&  {\renewcommand{\arraystretch}{0.8} \begin{tabular}[c]{@{}c@{}}\textbf{0.879}\\ {\scriptsize (0.0014)}\end{tabular}}       
&  {\renewcommand{\arraystretch}{0.8} \begin{tabular}[c]{@{}c@{}}\textbf{0.809}\\ {\scriptsize (0.0028)}\end{tabular}}        
&  {\renewcommand{\arraystretch}{0.8} \begin{tabular}[c]{@{}c@{}}\textbf{0.837}\\ {\scriptsize (0.0006)}\end{tabular}}            
&  {\renewcommand{\arraystretch}{0.8} \begin{tabular}[c]{@{}c@{}}\textbf{0.884}\\ {\scriptsize (0.0000)}\end{tabular}}       
&  {\renewcommand{\arraystretch}{0.8} \begin{tabular}[c]{@{}c@{}}\textbf{0.757}\\ {\scriptsize (0.0017)}\end{tabular}}       
&  {\renewcommand{\arraystretch}{0.8} \begin{tabular}[c]{@{}c@{}}\textbf{0.844}\\ {\scriptsize (0.0015)}\end{tabular}}            
&  {\renewcommand{\arraystretch}{0.8} \begin{tabular}[c]{@{}c@{}}\textbf{0.892}\\ {\scriptsize (0.0009)}\end{tabular}}       
&  {\renewcommand{\arraystretch}{0.8} \begin{tabular}[c]{@{}c@{}}\textbf{0.781}\\ {\scriptsize (0.0044)}\end{tabular}}
\\ \bottomrule

\end{tabular}

    \vspace{-0.5cm}
\end{table*}

\subsection{Analysis of Training Objective}
\label{sec:theo_analysis}

While contrastive learning has been widely explored in other NLP tasks~\citep{simcse,gunel2020supervised,su2021tacl,wu2020clear}, there lacks an effort for story discovery. We analyze how the learning objective in Equation \ref{eq:loss} leads to an effective embedding space for story discovery in terms of the alignment and uniformity\,\citep{align_uniform}; The former measures how close articles of the same story are while the latter measures how uniformly distributed random articles are. Equations \ref{eq:align} and \ref{eq:uniform} formalize the two properties with an embedding function $f(\cdot)$\,\citep{align_uniform}.

\begin{equation}
\small
\label{eq:align}
    l_{\text{align}} \triangleq 
    \underset{(x,x^+) \sim p_{pos}}{\mathbb{E}} \| f(x) - f(x^+) \|^2 \text{ and}
\end{equation}
\vspace{-0.3cm}
\begin{equation}
\small
\label{eq:uniform}
    l_{\text{uniform}} \triangleq \log \underset{(x,y) \underaccent{\sim}{i.i.d.} p_{data}}{\mathbb{E}} e^{-2 \| f(x) - f(y) \|^2},
\end{equation}


The contrastive loss in Equation \ref{eq:loss} can be rewritten as follows with the article distribution $\mathbb{D}$ and the corresponding stories $\mathbb{S}$,
\begin{equation}
\small
\label{eq:loss_analysis1}
    \mathcal{L}_{cts} \sim -\underset{d \sim \mathbb{D}}{\mathbb{E}} \Big( \text{log}\frac{e^{\text{cos}({R_{d},R_{S^{d}})}/\tau}}{\sum_{S \in \mathbb{S}}e^{\text{cos}({R_{d},R_{S})}/\tau}} \Big) \quad\quad\quad\quad\quad\quad\quad
\end{equation}
\vspace{-0.3cm}
\begin{equation}
\small
\label{eq:loss_analysis2}
\begin{split}
    \quad\quad = -\underset{d \sim \mathbb{D}}{\mathbb{E}} \Big(&\text{cos}(R_{d},R_{S^{d}})/\tau-\text{log}\sum_{S \in \mathbb{S}} e^{\text{cos}(R_{d},R_{S})/\tau} \Big).
\end{split}
\end{equation}
Since a story representation $R_{S^d}$ of a story $S^d$ of an article $d$ is the mean representation of articles in the same story\,($d^+$), i.e., $R_{S^d} = \underset{d^+ \sim S^d}{\mathbb{E}} R_{d^+}$, the cosine similarity between an article and a story in each term in Equation \ref{eq:loss_analysis2} is rewritten to:
{\small
\begin{align}
\label{eq:loss_analysis3}
    \text{cos}(R_{d},R_{S^{d}}) \sim \underset{d^+ \sim S^d}{\mathbb{E}} \text{cos}(R_d,R_{d^+}) \\
    \label{eq:loss_analysis4}
    \text{cos}(R_{d},R_{S}) \sim \underset{d' \sim S}{\mathbb{E}} \text{cos}(R_d,R_{d'}).
\end{align}
}

Consolidating Equations \ref{eq:loss_analysis2} to \ref{eq:loss_analysis4} by assuming the number of negative articles ($d^-$ in different stories with the story of $d$) approaches infinity, following the relevant work\,\cite{align_uniform, simcse}, gives
\begin{equation}
\small
\label{eq:loss_analysis5}
\begin{split}
    \mathcal{L}_{cts} \sim -\underset{d \sim \mathbb{D}}{\mathbb{E}} \Big(&\underset{d^+ \sim S^d}{\mathbb{E}} \text{cos}(R_d,R_{d^+})/\tau -\text{log}\sum_{S \in \mathbb{S}} e^{\underset{d' \sim S}{\mathbb{E}} \text{cos}(R_d,R_{d'})/\tau} \Big) \\
\end{split}
\end{equation}
\vspace{-0.3cm}
\begin{equation}
\small
\label{eq:loss_analysis6}
\begin{split}
    \quad\quad = &-\frac{1}{\tau}\underset{(d,d^+) \sim S^d}{\mathbb{E}} \text{cos}(R_{d},R_{d^+})+\underset{d \sim \mathbb{D}}{\mathbb{E}} \Big( \text{log}  \underset{d^- \sim \mathbb{D}}{\mathbb{E}} e^{\text{cos}(R_{d},R_{d^-})/\tau} \Big).
\end{split}
\end{equation}
Minimizing the two terms in Equation \ref{eq:loss_analysis6} promotes the alignment (by the first term) and the uniformity (by the second term).




\section{Experiments}
We evaluate the efficacy and efficiency of \algname{} for unsupervised online story discovery to answer the following questions.

\begin{itemize}[leftmargin=12pt, noitemsep]
\item How \emph{accurately} does \algname{} discover true stories over real article streams? (Section \ref{exp:accuracy}).
\item Are the techniques employed in \algname{} \emph{effective}? How good is the resulting \emph{embedding space}? How \emph{sensitive} and \emph{scalable} is \algname{} to various evaluation settings? (Section \ref{exp:analysis}).
\item How effectively does \algname{} identify \emph{recent real stories} and the story-relevant parts in the articles? (Section \ref{exp:casestudy}).
\end{itemize}

\subsection{Experiments Setting}
\textbf{\uline{Data sets.} } We used three existing real news data sets with story labels: News14\,\citep{miranda} with 16,136 articles of 788 stories, WCEP18\,\citep{WCEP} with 47,038 articles of 828 stories, and WCEP19\,\citep{WCEP} with 29,931 articles of 519 stories. We also prepared CaseStudy with articles of four recent popular stories. Refer to Section \ref{apx:datasets} for details.

\noindent
\textbf{\uline{Compared algorithms.} } We evaluated four state-of-the-art existing algorithms which are applicable for \emph{unsupervised} and \emph{online} story discovery: ConStream\,\citep{constream}, NewsLens\,\citep{newslens}, BatClus\,\citep{miranda}, and DenSps\,\citep{staykovski},  For \algname{}, we used two popular PSEs respectively based on Sentence-BERT\,\citep{sentencebert} (referred to as \textbf{\algname{}+SBERT}, default) and Sentence-T5\,\citep{sentencet5} (referred to as \textbf{\algname{}+ST5}). For a more fair comparison, we also prepared the advanced variants of existing algorithms with PSEs (e.g., +ST5 and +SBERT). Refer to Section \ref{sec:related_work} and Section \ref{apx:algorithms} for more details. 

\noindent
\textbf{\uline{Evaluation metrics.} } We used B-cubed scores ($B^3$-P, $B^3$-R, and $B^3$-F1)\,\citep{b-cubed}, for evaluating article-wise accuracy for the homogeneity and the completeness of stories, in addition to two clustering quality measures: Adjusted Rand Index (ARI)\,\citep{ari} and Adjusted Mutual Information (AMI)\,\citep{ami}. For each metric, the average score over sliding windows in the entire article stream is reported. 

\subsection{Overall Story Discovery Performance}
\label{exp:accuracy}
Table \ref{tbl:overall_accuracy} shows $B^3$-F1, AMI, and ARI results (note that $B^3$-P and $B^3$-R scores are comprehensively considered in the $B^3$-F1 scores). It is notable that all the existing algorithms improved their scores with the help of PSEs (+SBERT was better than +ST5 in overall), indicating that the general embedding capability of PSEs is also useful in story discovery. Nonetheless, \algname{} achieved the highest scores by outperforming the best scores of the compared algorithms by 10.3\% in $B^3$-F1, 8.6\% in AMI, and 20.7\% in ARI when averaged over all data sets. This clearly demonstrates the merits of story-indicative adaptive modeling and the employed techniques for unsupervised story discovery. We further analyze \algname{}+SBERT (referred to as \algname{} for brevity) in the following sections. 

\begin{table}[!t]
    \centering
    \caption{$B^3$-F1 results of ablation study.}
    \vspace{-0.3cm}
    \label{tbl:ablation}
    \small
\setlength{\tabcolsep}{3pt}
\renewcommand{\arraystretch}{0.9}
\begin{tabular}{l|ccc}
\toprule
   &    News14          &   WCEP18     &    WCEP19   \\ \midrule
\rowcolor{lightgray}
\algname{} (default)   
& \textbf{0.872}
& \textbf{0.837}
& \textbf{0.844}
\\
\ w/o memory replay (MR)  & 0.865             & 0.821    &    0.825 \\
\ w/o prioritized-aug. (PA)    &  0.864           & 0.832       &  0.831      \\
\ w/o MR \& PA    &  0.862            &  0.814      &  0.820      \\
\ w/o story-indicative emb.  & 0.851             &   0.767    &  0.774      \\\midrule
\algname{} with MR from  &              &        &      \\
\ Last two windows   &   0.869            & 0.831       &  0.839    \\
\ Last three windows   &   0.868           & 0.834       &   0.838   \\
\ All previous windows &  0.871            &  0.835      &  0.842    \\ \midrule
\algname{} with PA by   &             &        &      \\
\ Attentive pooling weights  &  0.866            &    0.831 &  0.833  \\
\ Random sentence shuffle  &    0.869          &   0.831  & 0.838    \\
\ Random feature dropout   &   0.868     &  0.829   & 0.836   \\ \midrule
\algname{} updated only in   &             &        &      \\
\ The initial window  &   0.855           &  0.776       &  0.791  \\
\ Random 30\% windows  &    0.862          &   0.826  & 0.830    \\
\ Random 70\% windows  &   0.866     &  0.830   & 0.834   \\
\bottomrule
\end{tabular}

    \vspace{-0.5cm}
\end{table}

\subsection{Analysis of \algname{}}
\label{exp:analysis}
\textbf{\uline{Ablation study.} }
To evaluate the efficacy of each component deployed in \algname{}, we conducted ablation studies on confidence-aware memory replay (MR), prioritized-augmentation (PA), and story-indicative embedding. Table \ref{tbl:ablation} shows the $B^3$-F1 results while other metrics showed similar trends. The story-indicative embedding showed the highest impact on the performance (i.e., the performances without it were comparable to the best results of compared algorithms), demonstrating its merits. MR and PA also contributed to the performance improvement, while the default strategies employed in each technique were more effective than the other alternatives. Specifically, (1) MR from the wider temporal scope of an article stream was not always helpful since it fails to keep the latest context of article streams. While considering all the previous articles showed comparable performance to the best results, it will not be practical. (2) PA by the attentive pooling weights or random shuffle/dropout degraded the performances as they could not generate valuable samples on par with the ones generated by using MHA weights reflecting implicit contextual relationships between sentences. (3) Story-indicative embedding not continuously adapted to the entire stream led to suboptimal performances. \algname{} with the story-indicative article encoder updated only in the first window achieved slightly better performance than the one without the encoder. The performance increased when the encoder was updated with more windows and was maximized when continuously updated with all windows. This shows that continual adaptation of \algname{} is necessary to deal with evolving articles and stories.

\begin{figure}[!t]
    \centering
     \begin{subfigure}[b]{0.45\columnwidth}
         \centering
         \includegraphics[width=\textwidth]{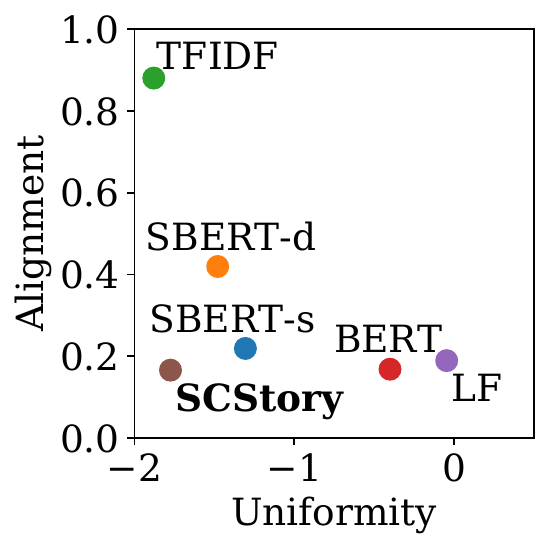}
         \vspace{-0.6cm}
         \caption{Alignment ($\downarrow$ is better) and uniformity ($\downarrow$ is better).}
         \label{fig:align_uniform}
     \end{subfigure}
     \hfill
     \begin{subfigure}[b]{0.51\columnwidth}
         \centering
         \includegraphics[width=\textwidth]{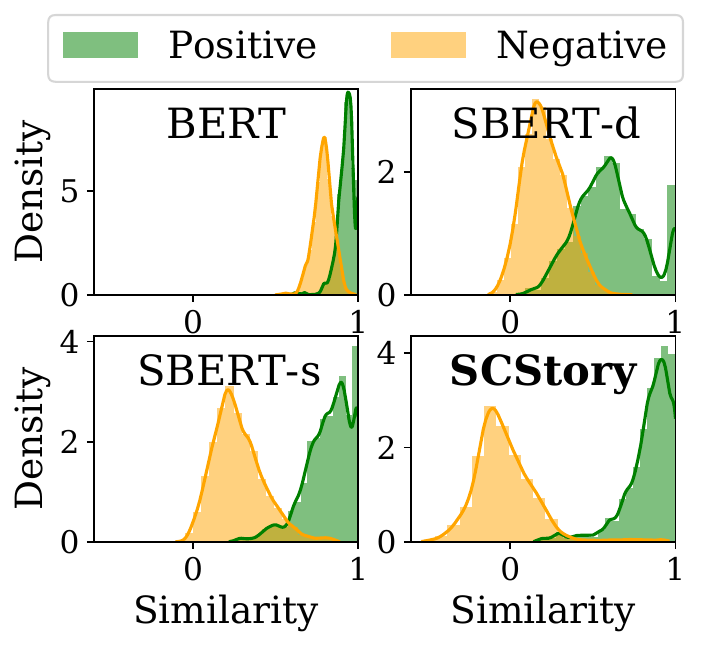}
         \vspace{-0.6cm}
         \caption{Positive ($\uparrow$ is better) and negative ($\downarrow$ is better) similarities.}
         \label{fig:pos_neg_sims}
     \end{subfigure}
    \vspace{-0.3cm}
    \caption{Comparison of different embedding strategies.}
    \label{fig:comparison}
    \vspace{-0.3cm}
\end{figure}

\begin{figure*}[!t]
\begin{subfigure}[b]{\columnwidth}
     \centering
     \includegraphics[width=\columnwidth]{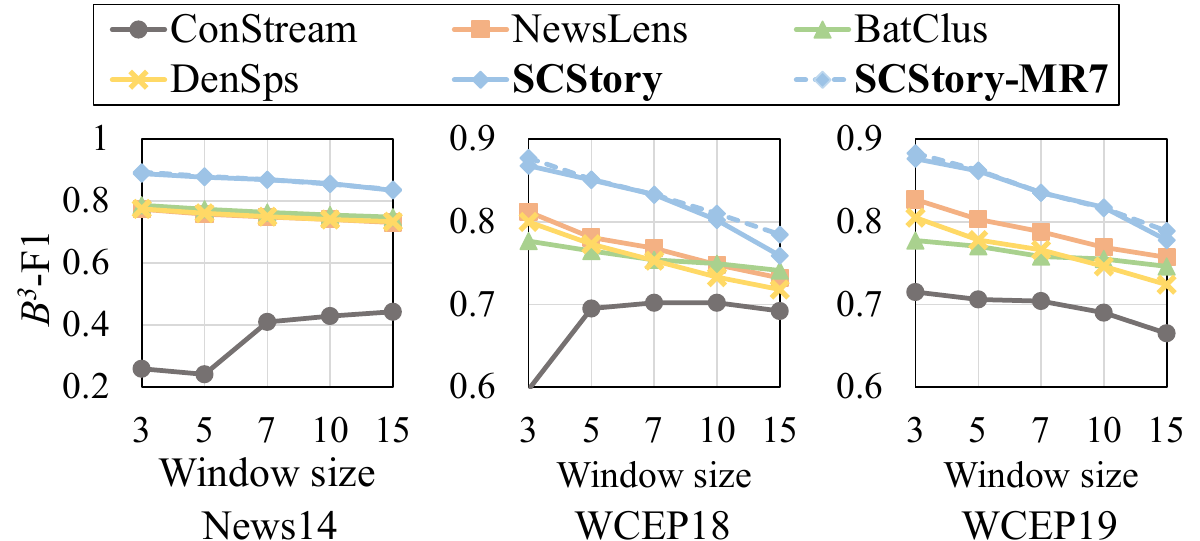}
     \vspace{-0.5cm}
    \caption{$B^3$-F1 scores.}
    \label{fig:varyingW_f1}
\end{subfigure}
\hfill
\begin{subfigure}[b]{\columnwidth}
     \centering
     \includegraphics[width=\columnwidth]{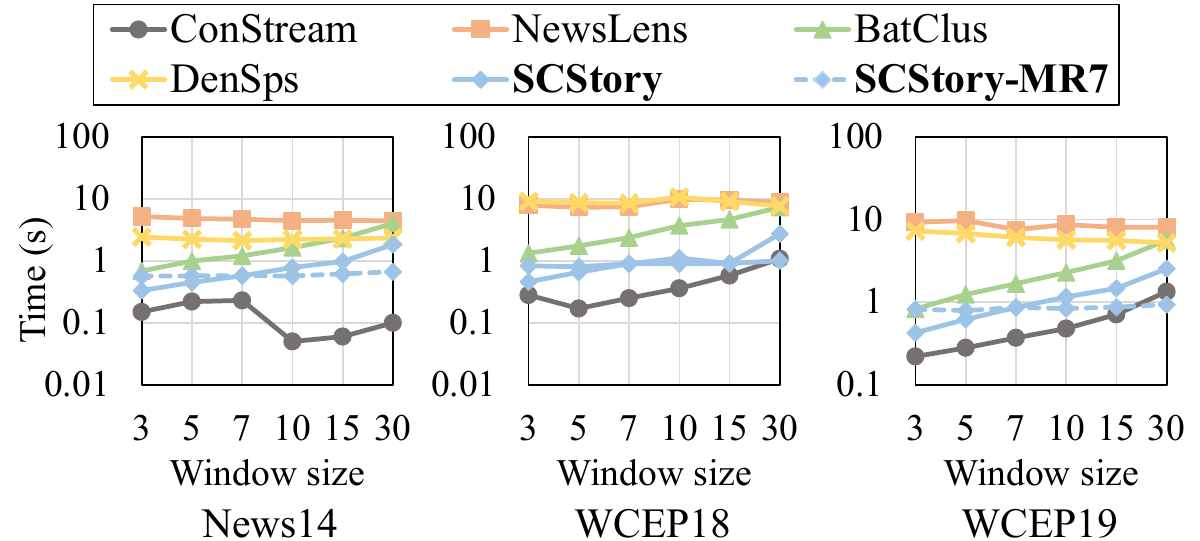}
     \vspace{-0.5cm}
    \caption{Running time (the y-axes are in a log scale).}
    \label{fig:varyingW_time}
\end{subfigure}
\vspace{-0.35cm}
\caption{Scalability study results with varying window sizes}
\vspace{-0.3cm}
\label{fig:scalability}
\end{figure*}

\begin{figure}[!t]
    \centering
     \begin{subfigure}[b]{0.49\columnwidth}
         \centering
         \includegraphics[width=\textwidth]{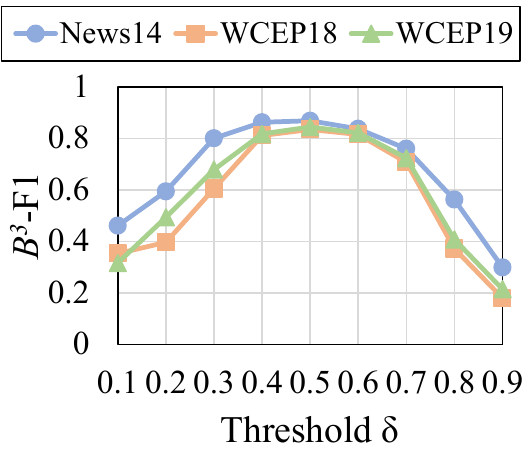}
         \vspace{-0.5cm}
         \caption{The threshold $\delta$.}
         \label{fig:varying_thred}
     \end{subfigure}
     \hfill
     \begin{subfigure}[b]{0.49\columnwidth}
         \centering
        \includegraphics[width=\textwidth]{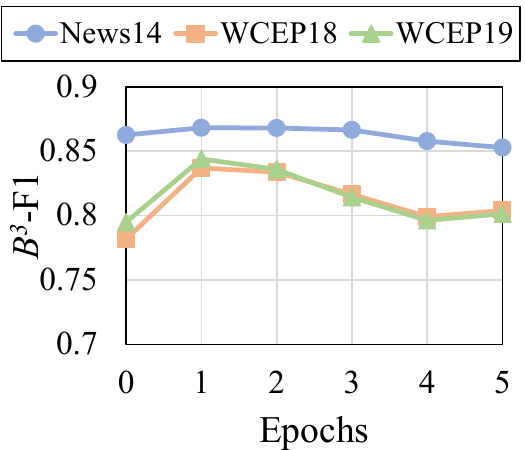}
         \vspace{-0.5cm}
         \caption{The number of epochs.}
         \label{fig:varying_epoch}
     \end{subfigure}
    \vspace{-0.3cm}
    \caption{$B^3$-F1 scores with varying hyperparameters.}
    \label{fig:hyperparameter}
    \vspace{-0.17cm}
\end{figure}


\noindent
\textbf{\uline{Embedding quality study.} }\label{exp:align_uniform}
We evaluated the embedding space created by \algname{} based on \emph{alignment} and \emph{uniformity}\,\citep{align_uniform}, discussed in Section \ref{sec:theo_analysis}. In addition, we measured the similarity distribution between positive articles and that between negative articles to verify how well the two distributions are separated. 

To compare these properties, we prepared various design alternatives in embedding articles: 1) \emph{TFIDF}\,(a representative sparse symbolic embedding); 2) \emph{BERT}\,(a mean pooling of final output tokens by BERT); 3) \emph{SBERT-d}\,(an output by Sentence-BERT from a whole article as an input); 4) \emph{SBERT-s}\,(a mean pooling of outputs by Sentence-BERT from individual sentences); 5) \emph{LF}\,(an output by Longformer\,\citep{longformer}, specifically designed for long sequences, from a whole article); and  6) \emph{\algname{}}\,(+SBERT). 

Figure \ref{fig:comparison} shows the comparison results in WCEP19. It is clear that SBERT-s balances the alignment and uniformity much better than the alternatives, which result in too specific (e.g., higher alignment from TFIDF or SBERT-d) or too general (e.g., higher uniformity from BERT or LF) representations, and also shows fairly well-separated similarity distributions; this demonstrates that sentence representations from a PSE can provide quality base representations to begin with. As expected, \algname{} further optimizes them for story-indicative article embedding and improves both the alignment and uniformity and the similarity distributions. 

\smallskip
\noindent
\textbf{\uline{Scalability study.} }
We evaluated the scalability of \algname{} and the existing algorithms in terms of $B^3$-F1 scores and the average running time of sliding windows, by varying the window sizes. A window size indicates the temporal scope of interest for current articles and stories in the window, which is the most critical factor affecting scalability in a streaming setting. All of the compared algorithms use Sentence-BERT as the base embedding model which showed the best performances. For \algname{}, we also prepared a variant of \algname{}, referred to as \algname{}-MR7, with the memory replay from the last 7 days of articles (i.e., the default window size) to investigate how the static memory replay affects the scalability.

As shown in Figure \ref{fig:varyingW_f1} and Figure \ref{fig:varyingW_time}, \algname{} (and \algname{}-MR7) consistently achieved the highest $B^3$-F1 scores for all cases while taking only less than or around a second for each sliding window, which was smaller than the most cases except ConStream (of which $B^3$-F1 scores were lowest, though). More specifically, inference of new articles and training of $\mathcal{M}$ respectively accounted for 13.5\% and 86.5\% of the running time, when averaged over all cases. This demonstrates the practical efficiency of \algname{} in an online scenario due to its lightweight inference and training mechanisms. 

Overall, a larger window size degrades the accuracy due to a wider temporal scope of stories to consider, while most of the real stories last only for a few days. The running time increased with the window size for ConStream and BatClus with more cluster management as well as \algname{} with more memory replay. The running time was almost consistent but much higher for NewsLens and DenSps with the incremental merging of communities on graph-based embedding. \algname{}-MR7 with the static memory replay, however, showed the almost consistent but much lower running time than them, while achieving the constantly higher $B^3$-F1 scores. This indicates that the appropriate scope of memory replay can help both the accuracy and the efficiency of \algname{}.

\noindent
\textbf{\uline{Hyperparameter study.} } Figure \ref{fig:hyperparameter} shows the effects of the threshold $\delta$ and the number of epochs, which are the two main hyperparameters respectively used to initialize a new story and to update $\mathcal{M}$ (Refer to Section \ref{apx:hyperparameter} for the results of the other hyperparameters). Conforming to the practical boundaries observed in the similarity distributions of SBERT-s and \algname{} in Figure \ref{fig:pos_neg_sims}, the value of 0.5 serves as the best threshold for initiating a new story. In the case of the number of epochs, updating $\mathcal{M}$ with the target temporal context (i.e., a current window) by only a single epoch was enough to keep the model up-to-date due to its lightweight structure, demonstrating the efficiency of \algname{}.

\subsection{Case Study}
\label{exp:casestudy}
\begin{figure*}
    \centering
    \includegraphics[width=\textwidth]{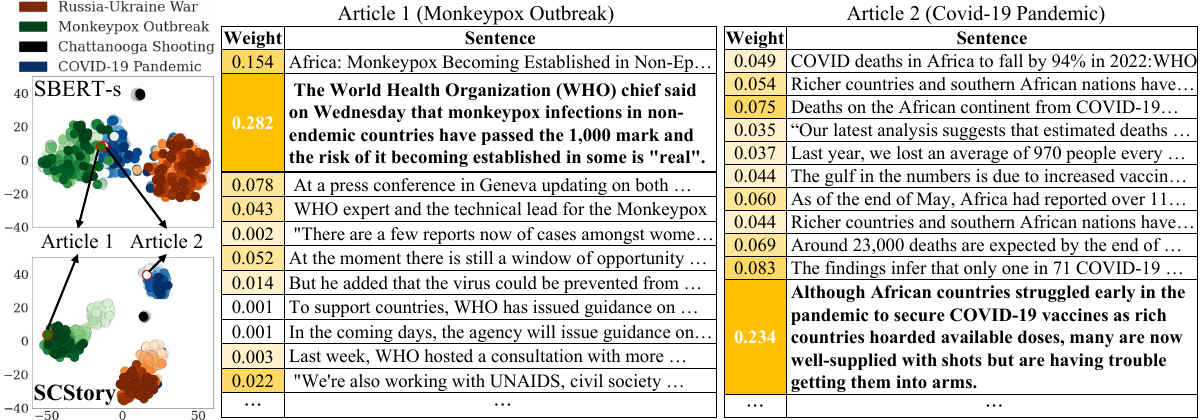}
    \vspace{-0.7cm}
    \caption{[Left] the article representations in CaseStudy by SBERT-s and \algname{} (darker markers indicate newer articles). [Right] the two example boundary articles and their sentences with attentive pooling weights learned by \algname{}.}
    \label{fig:casestudy}
    \vspace{-0.4cm}
\end{figure*}

We conducted qualitative analysis on CaseStudy constructed by collecting articles about four popular recent stories with conflicts and diseases: \texttt{Russia-Ukraine War}, \texttt{Monkeypox Outbreak}, \texttt{COVID-19 Pandemic}, and \texttt{Chattanooga Shooting}.
Figure \ref{fig:casestudy} shows the T-SNE\,\citep{tsne}-based 2D visualizations of the article representations embedded with SBERT-s (in the upper left part), that indicates the baseline embedding used for the variants of compared algorithms, and those embedded with \algname{} (in the lower left part). Note that the articles are color-coded according to their true story labels for convenience. While the boundaries between stories except for \texttt{Chattanooga Shooting} are not very clear in the embedding space with SBERT-s, due to the confusing articles with similar semantics (e.g., disease and death), \algname{} improved the embedding space for stories to be more clearly separated from one another. Specifically, while both of the two example boundary articles include similar descriptions about a disease, Africa, and WHO, \algname{} successfully discriminated the two articles by identifying how each sentence in the articles is relevant to its own story (as provided in the right part). Moreover, it is worth noting that \algname{} also effectively reflects the temporal changes of article representations in different stories, which can be further used to understand how stories evolve over time. Please refer to Section \ref{apx:casestudy} for a more detailed analysis with article representations for each story.

\section{Related Work}
\label{sec:related_work}

Mining news articles has been long studied through the earlier efforts in topic detection and tracking\,\citep{TDT} and offline (i.e., retrospective) story analysis\,\citep{retro1, retro4, retro3} to the more recent work for online (i.e., prospective) story discovery\,\cite{storyforest_conf, miranda, linger, saravan, constream, newslens, staykovski}, enabling various downstream tasks such as recommendation\,\cite{news_recom}, summarization\,\citep{story_summary}, and topic mining\,\cite{lee2022taxocom, topicexpan}.

Some work for online story discovery adopted a \emph{supervised} approach\,\citep{storyforest_conf, miranda, linger, saravan} by using external knowledge, such as given story labels and entity labels, to prepare the embedding and clustering models. For instance, Miranda et al.\,\citep{miranda} uses a labeled training set to learn the weights for various embedding and temporal similarities and a cluster assignment threshold. However, they use a static model trained offline and use the model only for inference purposes to news article streams. The external knowledge is not always available since it is not only expensive to acquire with human annotations but also hard to keep up-to-date. Thus, the pre-trained model can be easily outdated as it does not adapt to the evolving distributions of news articles from diverse stories. 

On the other hand, the \emph{unsupervised} work\,\citep{constream, newslens, staykovski} takes more practical approaches naturally applicable to evolving news article streams, which is also the scope of this work. They utilize symbolic features of articles for incremental clustering, such as overlapping keywords and TF-IDF scores which are readily available and updatable with new articles without relying on the external knowledge. Specifically, ConStream\,\citep{constream}, a popular streaming document clustering algorithm widely used for story discovery, exploits keywords-counts statistics and incremental clustering with micro-clusters. NewsLens\,\citep{newslens} uses keywords-based graph embedding, the Louvain community detection, and incremental community merging. Staykovski et al.\,\citep{staykovski} follows a similar graph-based approach with NewsLens but exploits TF-IDF- or doc2vec-based graph embedding. These methods, however, use explicit keywords statistics which are too short for capturing the diversified local context of article streams. A recent pretrained language model (e.g., Sentence-BERT\,\cite{sentencebert}) may advance the embedding quality and improve their performances (as we demonstrated in the evaluation of their variants in Section \ref{exp:accuracy}), but the indiscriminate embedding of the whole content is still too short for capturing the evolving global context of article streams. In this work, we utilize a generalized embedding capability of a pretrained language model while exploiting it specifically for story-indicative adaptive modeling of article streams with a self-supervised continual learning mechanism. 

Besides, some work has been proposed for online event detection and localization\,\cite{KPGNN, storyforest, triovecevent}, but they typically assume short messages (e.g., Twitter stream) and fine-grained event analysis. At the same time, some work introduced offline fine-tuning of customized language models for a news domain\,\citep{newsembed, newsbert} or continual pretraining of generalized language models for emerging corpora\,\citep{lifelongLM, continualLM}. They are orthogonal to our work in that their models can still be exploited further as a base encoder in our framework specifically for unsupervised online story discovery from article streams.

\vspace{-0.05cm} 
\section{Conclusion and Future Work}
We proposed a self-supervised continual learning framework \algname{} for unsupervised online story discovery. It realizes story-indicative adaptive modeling, fueled by confidence-aware memory replay and prioritized-augmentation. While existing algorithms utilize static and indiscriminate embedding, \algname{} automatically learns to identify the story-relevant information of articles and represents them to be more clearly clustered into stories. The proposed embedding strategy was theoretically analyzed and empirically evaluated with real news data streams and a case study to demonstrate its state-of-the-art performances. We believe this work could be a fundamental component for facilitating downstream tasks and applications. For instance, the identified story-indicative sentences can be further exploited to retrieve similar articles for recommendation or to get concise paragraphs for summarization.

We also discuss interesting directions to further improve \algname{} for future work. First, the definition of a story may need to be more relaxed for an article to belong to multiple stories. Then, \algname{} can be extended with post-refinement of discovered stories or soft/hierarchical clustering. Second, the passage of time can be more explicitly considered in a more flexible way to deal with stories over diverse temporal spans. Since simply using wider windows is neither effective nor efficient as discussed in the ablation study, \algname{} can manage a supplementary offline repository in conjunction with the sliding window. Lastly, weak supervision with story-relevant information (e.g., a topic or category in the article metadata) can be exploited as auxiliary guidance for \algname{} to boost its performance.

\clearpage
\begin{acks}
The first author was supported by Basic Science Research Program through the National Research Foundation of Korea (NRF) funded by the Ministry of Education (2021R1A6A3A14043765). The research was supported in part by US DARPA KAIROS Program No. FA8750-19-2-1004 and INCAS Program No. HR001121C0165, National Science Foundation IIS-19-56151, IIS-17-41317, and IIS 17-04532, and the Molecule Maker Lab Institute: An AI Research Institutes program supported by NSF under Award No. 2019897, and the Institute for Geospatial Understanding through an Integrative Discovery Environment (I-GUIDE) by NSF under Award No. 2118329. The views and conclusions contained in this paper are those of the authors and should not be interpreted as representing any funding agencies.
\end{acks}


\bibliographystyle{ACM-Reference-Format}

\balance
\bibliography{reference}

\nobalance
\appendix
\clearpage
\section{Supplementary Materials}
\label{apx:appendix}
\subsection{Time Complexity Analysis}
\label{apx:complexity}
\noindent
Let $N_{\mathbb{W}}$ and $N_{\mathbb{S}}$ be the number of articles and that of stories in a window, respectively, $N_{\mathcal{M}}$ be the model parameter size, $e$ be the number of epochs, and $b$ be the batch size. Then, (1) the time complexity for story-indicative embedding is $O(N_{\mathbb{W}}+N_{\mathbb{W}}N_{\mathcal{M}})$ where $O(N_{\mathbb{W}})$ is for initial sentence embedding (with a constant number of parameters from a frozen PSE) and $O(N_{\mathbb{W}}N_{\mathcal{M}})$ for article embedding. (2) The time complexity for story assignment is $O(N_{\mathbb{W}} + N_{\mathbb{W}}N_{\mathbb{S}})$ where $O(N_{\mathbb{W}})$ is for embedding stories and $O(N_{\mathbb{W}}N_{\mathbb{S}})$ is for assigning articles to stories. (3) The time complexity for self-supervised continual learning is $O(eb+ebN_{\mathbb{S}}N_{\mathcal{M}})$ where $O(eb)$ is for sampling and augmenting batches and $O(ebN_{\mathbb{S}}N_{\mathcal{M}})$ is for updating an encoder. Finally, since typically $N_{\mathbb{W}} \gg N_{\mathbb{S}}, e, b$, the total time complexity is $O(N_{\mathbb{W}}N_{\mathcal{M}})$. Note that $N_{\mathcal{M}}$ is 5.25M for \algname{}-SBERT and 2.95M for \algname{}-ST5 with default settings, which are much less than that of the typical language models (e.g., 110M for BERT\,\cite{bert}) owing to its lightweight architecture.

\subsection{Detailed Experiment Setting}
\textbf{\uline{Data sets}}
\label{apx:datasets}
The real news data sets used for simulating news streams to evaluate the compared algorithms are summarized in Table \ref{tbl:datasets}.
\vspace{-0.2cm}
\begin{table}[!h]
\caption{Data sets.}
\vspace{-0.45cm}
\small
\label{tbl:datasets}
\begin{tabular}{cccc}
\toprule
Data set    & \begin{tabular}[c]{@{}c@{}}\#Articles {\footnotesize (Avg \#Sentences)}\end{tabular}& \begin{tabular}[c]{@{}c@{}}\#Stroies {\footnotesize (Avg \#Articles/day)} \end{tabular}\\ \hline
\begin{tabular}[c]{@{}c@{}} News14\end{tabular} &  16,136 (21.4)                & 788 (8)       \\
\begin{tabular}[c]{@{}c@{}} WCEP18 \end{tabular}      &  47,038 (26.9)               & 828 (18)      \\
\begin{tabular}[c]{@{}c@{}} WCEP19 \end{tabular}    & 29,931 (27.6)                & 519 (18)      \\
\begin{tabular}[c]{@{}c@{}} CaseStudy \end{tabular}     &  1,068 (27.7)                & 4 (23)      \\
\bottomrule
\end{tabular}
\vspace{-0.2cm}
\end{table}

\begin{itemize}[leftmargin=10pt, noitemsep]
    \item \underline{News14}\,\citep{miranda}: a multilingual news data set which is widely used to evaluate online story discovery. We used English news articles published in 2014.
    \item \underline{WCEP}\,\citep{WCEP}: a large-scale news data set collected from Wikipedia Current Event Portal and Common Crawl Archive. We used articles in the stories of at least 50 articles and published in 2018 (referred to as WCEP18) and 2019 (referred to as WCEP19). 
    \item \underline{CaseStudy}: we collected the recent news articles through NewsAPI\footnote{\url{https://newsapi.org}} by using the queries with the titles of four popular stories about conflicts and diseases archived in Wikipedia Current Event Portal\footnote{\url{https://en.wikipedia.org/wiki/Portal:Current_events}}: 
    \texttt{Russia-Ukraine War}, \texttt{Monkeypox Outbreak}, \texttt{COVID-19 Pandemic}, and \texttt{Chattanooga Shooting}.
\end{itemize} 

 Articles in each data set contain a title, contents, a story label, and a publication date. An article stream is simulated with sliding windows of 7 days slid by a day from each data set by feeding articles in chronological order. All articles in each data set are used to evaluate the compared algorithms since we follow an unsupervised streaming setting (i.e., a prequential evaluation\,\citep{prequential} scheme for inferencing new articles first and then training if applicable).

\noindent
\textbf{\uline{Implementation of compared algorithms}}
\label{apx:algorithms}
We compared \algname{} with the four existing state-of-the-art algorithms that are designed for unsupervised online story discovery or can be used for the same purposes. We also prepared their variants with state-of-the-art PSEs.

\begin{itemize}[leftmargin=10pt, noitemsep]
    \item \underline{ConStream}\citep{constream}: For the micro-cluster assignment threshold, the standard score of previous article similarities is used. We tuned it between [0,3] to find the best value following the original work.
    
    \item \underline{NewsLens}\,\citep{newslens}: Following the original work, we tuned the number of overlapping keywords for creating an edge in a graph between [1,5] and the similarity threshold for merging communities between [0,1] to find the best value.
    
    \item \underline{BatClus}\footnote{\label{manual_name} We manually named the algorithms for brevity.}(by Miranda et al.\,\citep{miranda}\footnote{https://github.com/Priberam/news-clustering}): While it is originally designed for a supervised setting with given external knowledge, we adopted it in an unsupervised setting as it is also widely compared with unsupervised online story discovery algorithms. We set the weights for various similarities to ones, suggested by the original work, and tuned the similarity threshold for article-cluster assignment between [1,5] to find the best value, as the value around 3 was used through supervised training in the original work.
    
    \item \underline{DenSps}\cref{manual_name} (by Staykovski et al.\,\citep{staykovski}): As suggested by the original work, we used the TF-IDF-based embedding which showed better results than the doc2vec-based embedding. We tuned the similarity threshold for creating an edge between [0, 0.5] and that for merging clusters between [0.5, 1] to find the best values following the original work.
    
    \item \underline{\textbf{\algname{}}}: The proposed algorithm in this work. To update the story-indicative article encoder $\mathcal{M}$ in every sliding window, we used 1 epoch, a batch size $|\mathbb{B}|=256$ (128 each for sampling and augmenting), a temperature $\tau = 0.2$, the number of heads $n=4$, and the Adam optimizer\,\citep{ADAM} with a learning rate of 1e-5. The hidden and output dimensionalities $h_e$ and $h_c$ are set to the same dimensionality of the initial sentence representation by a PSE (e.g., 1,024 for Sentence-BERT and 768 for Sentence-T5). To decide whether to put a new article into an existing story or initiate a new story with the article, the threshold $\delta=0.5$ is used for the article-story assignment. The hyperparameter analysis is provided in Sections \ref{exp:analysis} and Section \ref{apx:hyperparameter}. 
\end{itemize} 

\noindent For \algname{} and the variants of existing algorithms, we used two popular PSE models: \emph{all-roberta-large-v1}\footnote{\label{sbert} \url{https://huggingface.co/sentence-transformers/all-roberta-large-v1}} for Sentence-BERT\,\cite{sentencebert} and \emph{sentence-t5-large}\footnote{\url{https://huggingface.co/sentence-transformers/sentence-t5-large}} for Sentence-T5\,\cite{sentencet5}, which show high sentence embedding performances and scalability\footnote{\url{https://www.sbert.net/docs/pretrained_models.html}}. Note that \algname{} employs the story-indicative encoder to exploit the initial sentence representations by a PSE further, while the variants of existing algorithms use the mean pooling of the initial sentence representations by a PSE as an article feature  (i.e., SBERT-s in the embedding quality study in Section \ref{exp:analysis}).

\noindent
\textbf{\uline{Computing platforms}}
All algorithms are implemented with Python 3.8.8 and evaluated on a Linux server with AMD EPYC 7502 32-Core CPU, 1TB RAM, and NVIDIA RTX A6000.

 \begin{figure*}[!t]
    \centering
     \begin{subfigure}[b]{\textwidth}
         \centering
         \includegraphics[width=0.95\textwidth]{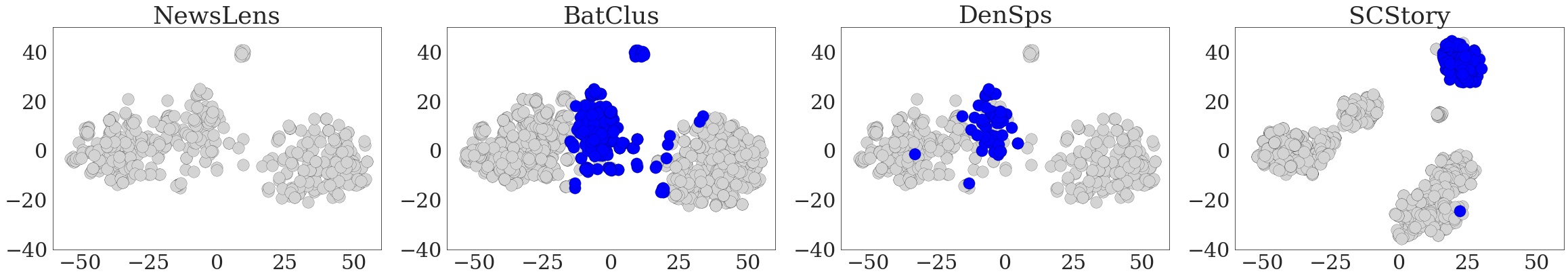}
         \caption{\texttt{COVID-19 Pandemic.}}
         \label{fig:covid19}
     \end{subfigure}
     \begin{subfigure}[b]{\textwidth}
         \centering
         \includegraphics[width=0.95\textwidth]{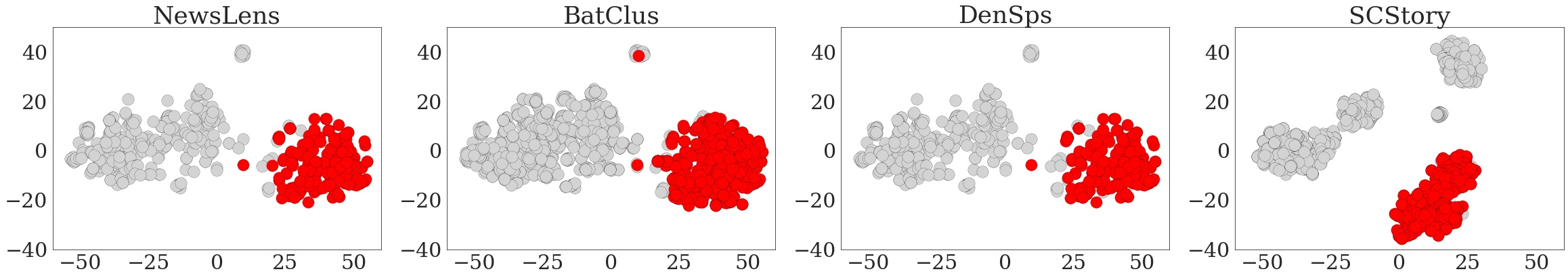}
         \caption{\texttt{Russia-Ukraine War.}}
         \label{fig:russia}
     \end{subfigure}
     \begin{subfigure}[b]{\textwidth}
         \centering
         \includegraphics[width=0.95\textwidth]{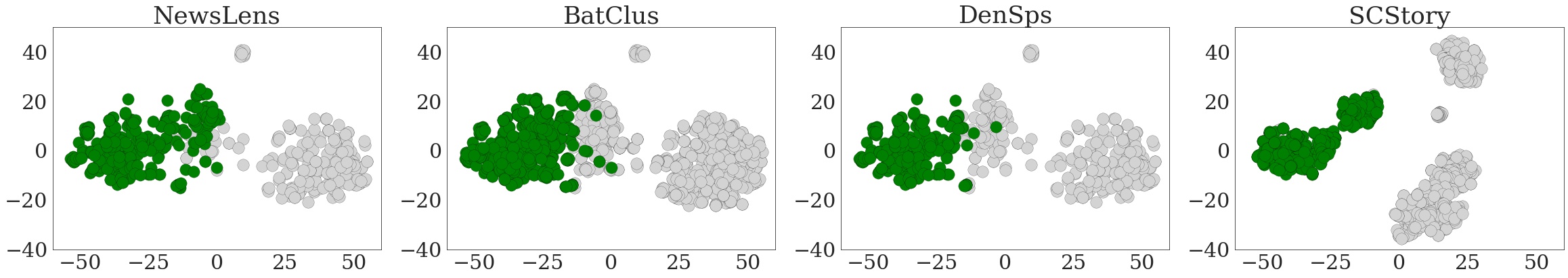}
         \caption{\texttt{Monkeypox Outbreak.}}
         \label{fig:monkeypox}
     \end{subfigure}
     \begin{subfigure}[b]{\textwidth}
         \centering
         \includegraphics[width=0.95\textwidth]{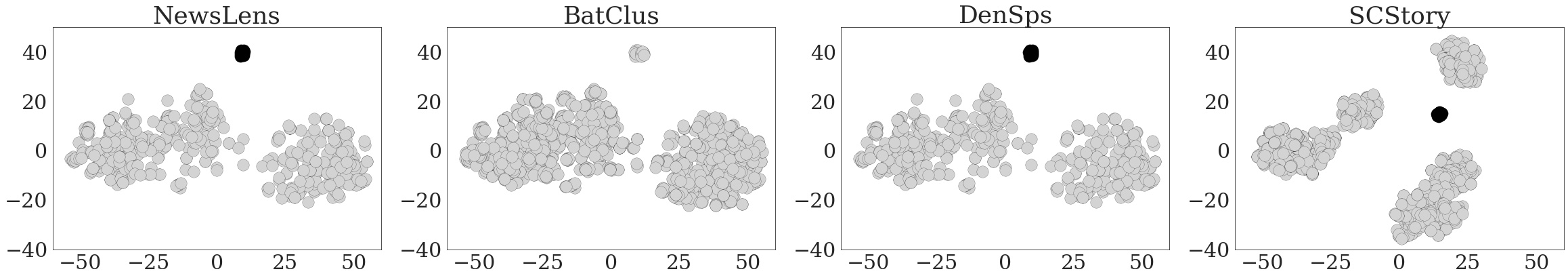}
         \caption{\texttt{Chattanooga Shooting.}}
         \label{fig:chattanooga}
     \end{subfigure}
    \vspace{-0.6cm}
    \caption{Visualization of article representations in the four stories discovered by each algorithm. Note that \algname{} learns its own embedding space while the compared algorithms use the embedding by SBERT-s.}
    \vspace{-0.1cm}
    \label{fig:casestudy_embeddings}
\end{figure*}

\noindent
\textbf{\uline{Pretrained models used in Section \ref{exp:analysis} }}
In the embedding quality study, BERT uses a pretrained model
bert-large-cased\footnote{\url{https://huggingface.co/tftransformers/bert-large-cased}}, SBERT-d and SBERT-s use a pretrained model 
all-roberta-large-v1\cref{sbert}, and LF uses a pretrained model
longformer-base-4096\footnote{\url{https://huggingface.co/allenai/longformer-base-4096}}.

\begin{figure}[!t]
    \centering
     \begin{subfigure}[b]{0.49\columnwidth}
         \centering
         \includegraphics[width=\textwidth]{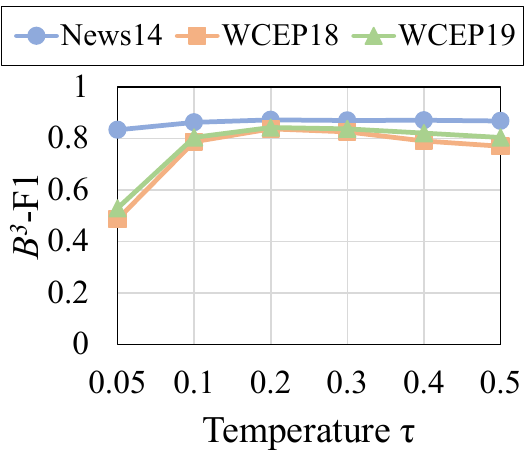}
         \vspace{-0.5cm}
         \caption{The temperature.}
         \label{fig:varying_temp}
     \end{subfigure}
     \hfill
     \begin{subfigure}[b]{0.49\columnwidth}
         \centering
         \includegraphics[width=\textwidth]{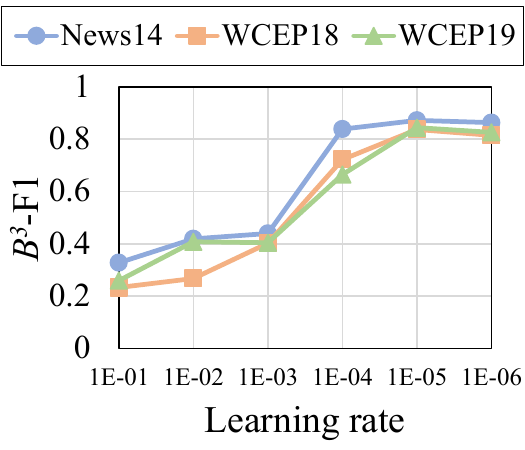}
         \vspace{-0.5cm}
         \caption{The learning rate.}
         \label{fig:varying_lr}
     \end{subfigure}
    \vspace{-0.3cm}
    \caption{$B^3$-F1 scores with varying hyperparameters.}
    \label{fig:hyperparameter2}
    \vspace{+0.14cm}
\end{figure}

\smallskip
\subsection{Detailed Analysis of the Case Study}
\label{apx:casestudy}
Figure \ref{fig:casestudy_embeddings} shows the stories discovered by \algname{} and the variants of three existing algorithms, where all of them use Sentence-BERT for the base embedding model (i.e., SBERT-s). Note that the discovered stories are matched to the true stories by the authors for convenience. While the variants of existing algorithms discovered incomplete and noisy stories or even failed to discover all stories (e.g., BatClus could not find \texttt{Chattanogga Shooting}), \algname{} successfully discovered the four stories almost similar to the true stories that are shown in Figure \ref{fig:casestudy}. The $B^3$-F1 score of \algname{} was 0.960, which was higher than those of the other algorithms (e.g., 0.858 for NewsLens, 0.902 for BatClus, and 0.871 for DenSps).

\subsection{Additional Hyperparameter Study Results}
\label{apx:hyperparameter}
Figure \ref{fig:hyperparameter2} shows the effects of the temperature $\tau$ in Equation \ref{eq:loss} and the learning rate for training $\mathcal{M}$, validating the default values ($\tau = 0.2$ and learning rate of 1e-5) used in \algname{}. The results of other metrics for $\tau$ and learning rate as well as the threshold $\delta$ and the number of epochs showed similar trends.

\end{document}